\documentclass[10pt]{article}
\usepackage{graphicx}
\usepackage{cogsci}
\usepackage{booktabs}
\usepackage{multirow}
\usepackage{subfigure}
\usepackage{amsmath}
\usepackage{amssymb}

\newcommand{\name}{{\sc TagBERT}}

\usepackage[
    backend=biber,
    style=apa,
    natbib=true,
    doi=false,
    isbn=false,
    url=false,
]{biblatex}
\usepackage{pslatex}
\usepackage{float} 

\title{Extracting Important Tokens in E-Commerce Queries with a Tag Interaction-Aware Transformer Model}
 
\author{
Md. Ahsanul Kabir$^{1}$ \quad Mohammad Al Hasan$^{1}$ \quad 
Aritra Mandal$^{2}$ \quad Liyang Hao$^{2}$ \quad Ishita Khan$^{2}$ \quad Daniel Tunkelang$^{2}$ \quad Zhe Wu$^{2}$ \\
$^{1}$Indiana University at Indianapolis, Indianapolis, Indiana, USA \\
$^{2}$eBay Inc., San Jose, California, USA \\
\texttt{\{mdkabir, alhasan\}@iu.edu} \quad
\texttt{\{arimandal, liyhao, ishikhan, dtunkelang, zwu1\}@ebay.com}
}
\cogscifinalcopy
\begin{document}

\maketitle

\begin{abstract}
The major task of any e-commerce search engine is to retrieve the most relevant inventory items, which best match 
the user intent reflected in a query. This task is non-trivial due to many reasons, including ambiguous queries, 
misaligned vocabulary between buyers, and sellers, over- or under-constrained queries by the presence of 
too many or too few tokens. To address these challenges, query reformulation is used, which modifies a
user query through token dropping, replacement or expansion, with the objective to bridge semantic gap between query tokens and users' search intent. Early methods of query reformulation mostly used statistical measures derived from 
token co-occurrence frequencies from selective user sessions having clicks or purchases. In recent years, supervised 
deep learning approaches, specifically transformer-based neural language models, or sequence-to-sequence models are 
being used for query reformulation task. However, these models do not utilize the semantic tags of a query token, which
are significant for capturing user intent of an e-commerce query. In this work, we pose query reformulation as 
a token classification task, and solve this task by designing a dependency-aware transformer-based language model, 
TagBERT, which makes use of semantic tags of a token for learning superior query phrase embedding. Experiments on large, real-life e-commerce datasets show that TagBERT exhibits superior performance than plethora of competing models, 
including BERT, eBERT, and Sequence-to-Sequence transformer model for important token classification task.
\end{abstract}

\section{Introduction}

For a satisfactory online shopping experience on e-commerce platforms like eBay, Alibaba, Amazon, etc. choosing a proper search query is of paramount importance. A well-formulated search query enables the search engine to retrieve the most
relevant inventory items, which best match the user intent. Unfortunately, for many items, formulating the most
important query is not easy due to many reasons, including, but not limited to, token ambiguity, misaligned vocabulary between buyers and sellers, over- or under-constrained queries by the presence of  too many or too few tokens. To 
overcome this, query reformulation~\cite{hirsch2020query,hasan2011query} is used which modifies a user query through token dropping, replacement or expansion. 
Replacement or expansion of tokens often lead to unintended consequences, so the most common reformulation trick is to
drop {\em non-important} tokens from a query. By dropping irrelevant or non-important tokens from a query, the search 
engine is able to retrieve more results with higher relevance score. Token dropping is particularly beneficial for tail
queries, which more often suffer from null and low recall~\cite{singh2011user,ordsim_mdkabir}. Besides better retrieval
experience, identifying important tokens also helps in developing better solution to several other platform services,
including ``related searches'', ``query expansion'', ``auto complete'', and ``improve spelling''. To facilitate 
on-the-fly token dropping in a production setting, a trained supervised model is needed, which can  differentiate between
important and non-important tokens of a query.

In existing literature, to our surprise, extraction of important tokens from e-commerce queries did not get much
attention, though many related works exist on e-commerce query engineering. There are many works which consider
the task of query similarity prediction~\cite{ordsim_mdkabir}, and query-item similarity prediction~\cite{kabir2023survey, bell2018title, hu2018reinforcement}. At the token level, there also exist a large number of works focusing on similar token identification for the purpose of query 
rewriting~\cite{Mandal2019QueryRU, qwrlattest}. Query expansion is also considered either by utilizing 
pseudo-relevance feedback from the top results retrieved from a first round of retrieval~\cite{pseudo}, or by
leveraging users' search logs~\cite{cui2002,qs2006}. Hirsch et al.~\cite{Hirsch2020} provide an in-depth analysis of
various query reformulation strategies, along with a model to predict whether a query in a session will be reformulated.
With the increasing popularity of neural language model, such as, BERT~\cite{bert:Embedding}, GPT-3~\cite{floridi2020gpt}, a recent 
work~\cite{liyang:2023} consider query reformulation as a  sequence to sequence prediction task using a variant of 
BERT model. Closest to our work is a work by Manchanda et al.~\cite{Manchanda:2019:term-weighting}, who uses historical
query reformulation logs to learn a weight on each token by using a distant-supervision approach. From the learned weight
one can easily distinguish between important and non-important tokens in a query.


Importance of a token in an e-commerce query largely depends on the role that a token plays within the context of entire
query phrase. Generally, an e-commerce query refers to a product where a token may refer to the name entity of the 
product or an attribute of the product. For example, for a query like ``nike women soccer jersey small'', the name 
entity is {\em soccer jersey}, and the attributes and values are {\em brand: nike}, {\em gender: women}, and 
{\em size: small}. Often the attributes are called structural annotations, or tags. In this example, brand, gender,
and size are three different tags. Tags of a token can be obtained from a product catalogue, or they can be obtained
automatically through an in-house token tag prediction model. Depending on the query, tokens belonging to some tags are more important than other.
Also, there are pair-wise dependency between a pair of tags in a token, as such, presence or absence of a token belonging 
to a given tag may render another token belonging to a different tag more or less useful. So tag information can be leveraged to identify importance of a token in a query, which is largely ignored in most of the existing works on token
importance. Multiple recent work~\cite{Wang2021QUEENNR, Farzana:2023:neuralqueryrewrite} have utilized token tags for query rewriting by modeling it as a sequence to sequence prediction task using transformer. We leverage 
tags for predicting important tokens in a query, by using a transformer based model, which we discuss next.

In this paper, we propose \name, a transformer-based model for extracting important tokens in e-commerce queries. \name\
follows a BERT like encoding of the query phrase for solving the token classification task. but unlike BERT, \name\
also generates an additional query phrase embedding by considering multi-head self-attention over token-pairs that are related through corresponding tag-associations. To obtain tag-association based embedding, \name\ first generates a 
tag dependency graph, so that self-attention mechanism exercises attention only over the tokens that are related through
the edges in the tag-dependency graph. Then the BERT-based phrase embedding and tag-dependency aware phrase
embedding are merged through a gating mechanism to obtain final query phrase embedding, which is passed through a
feed-forward layer for solving the token classification task. Our experiments over a large real-life e-commerce dataset
show that \name\ yield much superior results over a traditional BERT. It is also superior than an encoder-decoder based
sequence-to-sequence model built on a transformer model.
We claim the following contributions:
\begin{itemize}
    \item We introduce two graph learning approaches based on the tag-tag interaction. The first among them uses frequent itemset mining over a collection of representative queryset to generate a tag interaction graph, which is then incorporated into our model. The second method automatically learns the tag interaction graph in a dynamic fashion through attention mechanism as it incorporates the tag interaction into the model.
    
    \item We propose \name, a tag dependency-aware transformer model, which utilizes tag-dependency graph for obtaining query phrase embedding, which is superior that traditional BERT transformer based embedding for solving token importance classification task.
    \item We desing a novel gating mechanism in \name, which can automatically adjust the gating probability between traditional BERT embedding and tag-dependency aware BERT embedding depending on the quality of information in the tag-dependency graph.
    \item We perform extensive experiment over a large real-life e-commerce dataset to validate the effectiveness of \name. Our experiments show that \name\ outperforms a traditional BERT and an encoder-decoder based transformer model
    for solving the important token classification task.
\end{itemize}



\section{Methodology}\label{definition} 

In this section, we first provide a formal discussion of the important token classification task in the context of e-commerce queries. Then we provide the motivation and an overall framework of \name, our proposed model. Finally, we
provide an architecture details of \name.


\subsection{Problem formulation} 
We start by considering a source query denoted as $S$ and a destination query as $Q$. We apply pre-processing to both queries resulting in two sub-sequences for $S$, $s_1$, $s_2$ ... $s_M$, and $D$, $d_1$, $d_2$ ... $d_N$, where $M$ and $N$ represent the number of words in $S$ and $Q$ respectively after segmentation. As $Q$ is created by dropping some tokens from $S$, it follows that $M$ is greater than or equal to $N$. In addition, we generate the associated tags for $S$ as another sequence, denoted as $\tau$ = $t_1$, $t_2$ ...$t_M$, which are generated using eBay's query understanding pipeline that falls beyond the scope of our paper.
Moreover, we present the labels of the tokens of $S$ as a sequence $L$ denoted by $l_1$, $l_2$ ... $l_M$. For this particular token dropping task, we set number of labels, $K$ to 3, and $l_i$ is assigned a value based on the following conditions.

\[
    l_i= 
\begin{cases}
    1,& \text{if } \text{$s_i$ is a special token}\\
    2,& \text{if } \text{$s_i$ is not dropped}\\
    3 & \text{otherwise}
\end{cases}
\]

Additionally, we transform $l_i$ into a one-hot encoding vector of size $K$, denoted as $c_i$, using the one-hot-encoding method~\cite{harris2010digital,brownlee2017one}. Suppose there is a model, $\theta$, which predicts $p_1$, $p_2$ ... $p_M$, where $p_i$ represents the probability values predicted by the model for $l_i$. The objective of $\theta$ is to minimize the following function denoted by $\mathcal{L}$

\[\mathcal{L} = -\frac{1}{M}\sum_{i=1}^{M}\sum_{j=1}^{K}c_{i,j} \log p_{i,j}\]

\subsection{\name: Motivation and Design Justification}
Tags are structural annotation of a token in a query phrase, which refer to an item name or an attribute of the item.
Since, items of certain kind have a fixed set of attributes, distribution of tag co-occurrences in a query is not random, rather they are highly biased depending on the item category (a hierarchical organization of items into different groups). For example, in {\em Dress} or {\em Shoes} category, a query generally have the following tags: gender, size, color, and brand. On the other hand, a query in {\em Motor} category have tags like build-year, make, and model. Also, depending on a category, having a certain tag in a query, make it more or less likely to have another tag. If we have a model tag for a 
vehicle it is more likely to see a build-year tag in the same query. On the other hand, if we have a model tag in an
electronics query, it is less likely to see a year tag. Utilizing such insight is crucial for understanding which of the
tokens are important for a query.

To make use of tag association for the token classification task, we build an undirected tag association graph,
in which interacting pair of tags have an edge between them. The motivation for building the tag association graph is as follows: in a transformer architecture, a
token receive attentions from all other tokens. However, through tag association graph, we are able to prioritize the attention propagation only between tokens who are connected through an edge in the tag association graph. In other
words, \name\ brings a flavor of graph attention network~\cite{velivckovic2017graph} in its attention propagation mechanism to learn a  better embedding of the query phrase.

In \name, we actually learn two sets of embedding vectors for the query token; the first set of vectors are obtained through BERT, which allows pair-wise attention to propagate between any pair of tokens in a query. The second set is through tag-dependency aware
BERT, were attention propagation is allowed only along the edges of the tag-association graph. This redundancy is needed
to ensure that \name\ has the option to resort to standard BERT embedding through a gating mechanism, in case the 
tag-dependency BERT yield inferior quality embedding. This can happen for two reasons: first, the tag label of a token 
can be erroneous, and second, false-positive edges are added in the tag association graph due to poor statistics. This 
specifically happens for long tail queries, for which both the token labels and the tag association edges may be of poor 
quality due to data sparsity. 

In \name, we design our task as a token classification task. So, we only consider the encoding mechanism of a 
transformer, although transformers like BERT, GPT have encoder-decoder for sequence-to-sequence model. One can take
the source and destination of query phrase distinctly and consider the important token identification as a sequence
to sequence generation task. In experiment section, we compare with such a model and show that \name\ performs better 
than a traditional sequence-to-sequence model.

\begin{figure*}[t]
    \centering
    \includegraphics[scale=0.6]{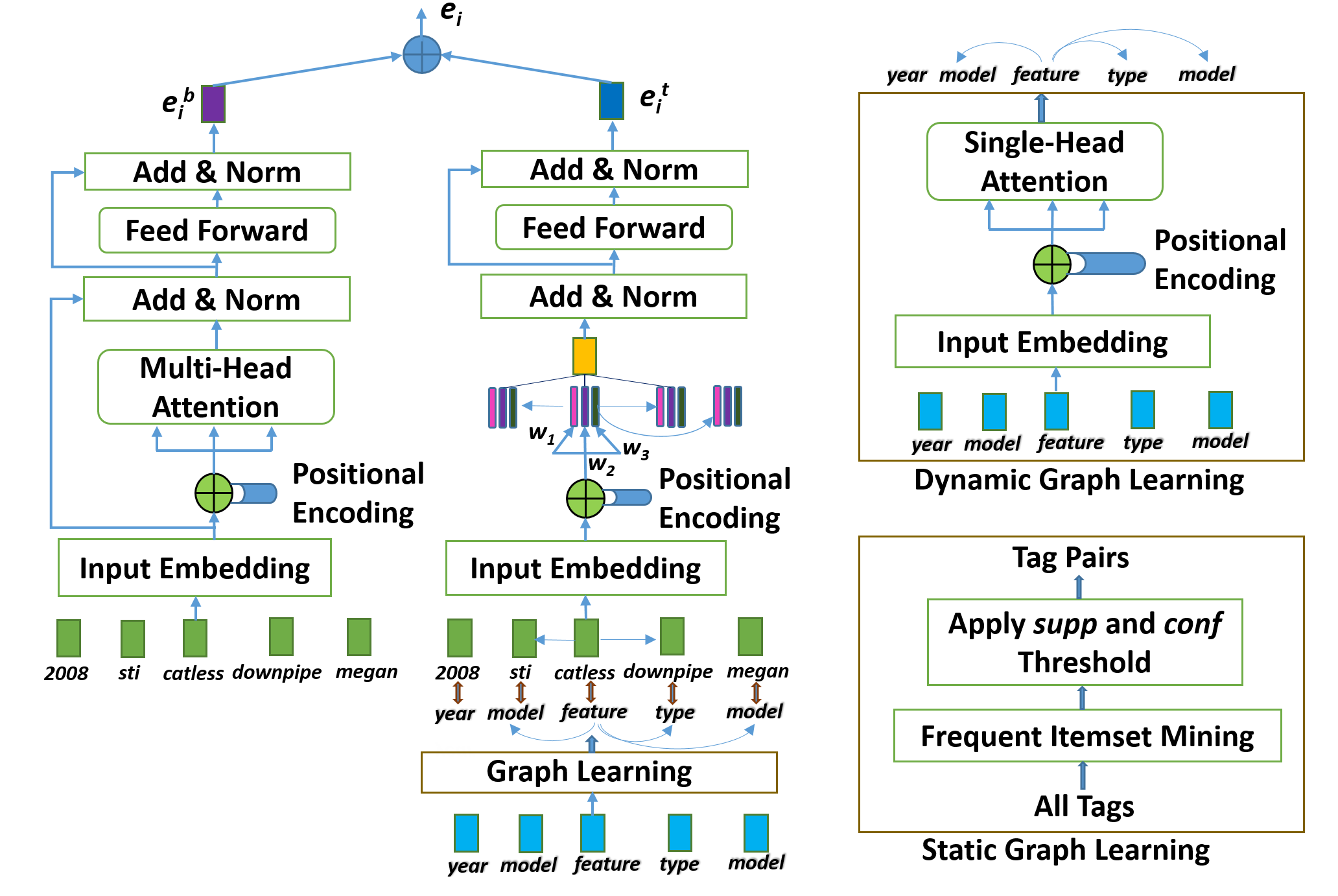}
    \caption{\name\ Model Architecture for Token Classification}
    \label{fig:classifier}
\end{figure*}

\subsection{Model Architecture} \label{model}
Figure~\ref{fig:classifier} depicts the \name\ model designed for token classification. The token embedding encoder of a traditional BERT~\cite{bert:Embedding} is shown on the left side of the model diagram, while the encoder~\cite{raganato2018analysis} architecture with a modified self-attention mechanism which incorporates a dependency-aware graph is on the right side of the model diagram. To facilitate the discussion, we use a running example query from the {\em motor} category, ``2008 sti catless downpipe megan''. At the bottom of the right part of the model architecture, we show the tag association example graph of this query. As we can see for this query, the tags are year, model, feature, type, and model, from left to right. The tag-association graph indicates edges between certain tag pairs, such as \textit{(feature, model)}, \textit{(feature, type)} which correspond to token pairs like \textit{(catless, megan)}, \textit{(catless, downpipe)}, and \textit{(catless, sti)}. The tag-association graph can be created as either a static graph or a dynamic graph, depending on the method used to learn the tag-tag interaction. In this research, we utilize both methods independently to create the graph, and
refer the model either as \name (Static), or \name (Dynamic), depending on the kind of graph used.

Left side of the model, which follows BERT identically takes all the tokens of $S$ as input except for some experiments where we modify the existing tokenizer to incorporate the tags for input embedding. However, in our current demonstration, we present the simplest architecture, where each token is passed through an initial trainable embedding layer for token ID and token type ID. Subsequently, positional encoding is added, and the resulting vector is passed through a multi-head attention layer. This multi-head attention layer learns attention for all tokens, and its output is calculated based on the attention scores of each token. The resulting output is then normalized using layer normalization and passed through a feed-forward layer to obtain $e_i^b$, which represents the output embedding for token $s_i$ from the left part of the model.

Meanwhile, the generated tree for $S$ is fit to the right side of the diagram. Like the BERT encoder architecture, embedding of input ID, and token type ID, and positional ID are gathered. These three embeddings are then added to construct a single vector $v_i$. The embedding of other tokens are also calculated in the similar fashion. Let the tokens connected with $s_i$ through edges is the set $\mathcal{V}$, and $s_k$ be any token in $\mathcal{V}$. Additionally, let there be three trainable matrices $\mathbf{W}_1, \mathbf{W}_2$, and $\mathbf{W}_3$. $v_i\mathbf{W}_1$, $v_i\mathbf{W}_2$, and $v_i\mathbf{w}_3$ are then query, key, and value vectors respectively for the token $s_i$. The affinity score of two connected tokens $s_i$, and $s_k$ is then calculated by the following equation.
\[a_{ik} = (v_i\mathbf{W}_1) * (v_k\mathbf{W}_2)^T\]
The attention value, $\alpha_{ik} $ (a scalar) is the softmax-score of these affinity values for all the tokens in $\mathcal{V}$ which actually represents how important the connected token is with respect to current token.

\[\alpha_{ik} = \frac{e ^ {a_{ik}}}{\sum_{j\in[1,|\mathcal{V}|]} e ^ {a_{ij}}}\]

\noindent The attention scores are used for attention based weighted sum for the output vector, $o_i$ from attention layer,
as shown in the next equation ($\mathbf{W}_4$ is another trainable matrix and $v_j\mathbf{W}_3$ is a value vector for the corresponding token.). Unlike BERT, $o_i$ is calculated for only those tokens which are connected to $s_i$. 

\[o_i = \sum_{k\in[1,|\mathcal{V}|]} \alpha_{ik} (v_j\mathbf{W}_3) \mathbf{W}_4\]

\noindent  $o_i$ is then passed through a normalization layer, and the output from \textbf{Add and Norm} layer, $\bar{o_i}$ is calculated using the following equation where $\gamma$, $\beta$ are trainable scalers, $\epsilon$ is a very small scaler constant, $\mu_i$, and $\sigma^2$ are the mean, and variance of the vector $o_i$.
\[\bar{o_i} = v_i + \gamma \odot \frac{o_i - \mu_i}{\sigma^2+\epsilon} + \beta\]
The output $\bar{o_i}$ is furthermore passed through a feed forward layer with GELU activation\cite{hendrycks2016gaussian} using another trainable matrix $\mathbf{W}_5$, and bias $b$.
\[FFN(\bar{o_i}) = GELU(\bar{o_i} \mathbf{W}_5 + b)\]
Additionally, $e_i^t$ is calculated passing the feed forward layer's output to another \textbf{Add and Norm} layer. Meanwhile, $e_i^b$ and $e_i^t$ are passed through another gate to form $e_i$ which is the token embedding of $s_i$ using
\name. This embedding is passed to another neural network for the token classification task.

\[e_i^s = \sigma(e_i^b\mathbf{W}_6 + c)\]
\[e_i = e_i^s \odot e_i^b + (1-e_i^s) \odot e_i^t\]

\noindent
{\bf \name (Dynamic):} To build tag-tag interaction dynamically, we use an attention-based architecture as shown on the top-right box of Figure~\ref{fig:classifier}. Output of this box is a dynamic graph, where each node is a tag-type (say, brand, color, size, etc.); edges between a pair of nodes are weighted with a probability value, denoting
the importance of tag-interaction between the pair of tags in the context of the given learning task.
If there are $\tau$ tokens in the query $S$, constructing the dynamic graph for $S$ requires $\tau*\tau$ probability values, all of which are learned through the tag-to-tag interaction of a deep learning model. In the \name\ architecture, we extend the existing dependency structure by incorporating a dynamic tag interaction graph learning mechanism. For this, we introduce all associated tags of the corresponding token to a probability learning architecture, wherein each tag has a unique ID and positional ID. These IDs are then passed to an embedding layer, generating word-piece representations denoted as $e_{i}^{tid}$ and $e_{i}^{p}$. Subsequently, both embeddings are summed to derive the representation of the tags.
\[e_{i}^{t} = e_{i}^{tid} + e_{i}^{p}\]

Note that this embedding is for the tags (say, size, or color) not for the tokens associated with that tag (say, if the color is blue, we are taking embedding of ``color'', not ``blue''). 
The tag embedding vectors then undergo a single head attention layer of BERT, which includes trainable matrices for query, key, and value vectors, determining the affinity scores between corresponding tags. To elaborate, let 
$\mathbf{W}_7$, $\mathbf{W}_8$, and $\mathbf{W}_9$ be the trainable matrices such that $e_{i}^{t} * \mathbf{W}_7$, $e_{i}^{t} * \mathbf{W}_8$, $e_{i}^{t} * \mathbf{W}_9$ be the corresponding key vector, query vector and value vector for 
the $i$'th tag. The affinity score between the $i$'th and $k$'th tag is then computed using the following equation, following a similar fashion as described earlier.

\[a_{ik}^t = (e_{i}^{t} * \mathbf{W}_7) * (e_{i}^{t} * \mathbf{W}_8)^T\]

\noindent The attention scores for the tags is the softmax values for all the tags in $S$ calculated by the following equation.

\[\alpha_{ik} ^t = \frac{e ^ {a_{ik} ^t}}{\sum_{j\in[1,\tau]} e ^ {a_{ij}^t}}\]
The calculated probability values and corresponding tag embeddings are retained for graph creation and subsequent concatenation, depending on the chosen model. It is important to emphasize that each token is assigned a tag, resulting in the requirement of $\tau*\tau$ probability values to determine the dynamic graph edge probabilities for \name.\\

\noindent
{\bf \name (Static)}: Tag-tag interaction graph can also be pre-computed (independently) from the training module) from frequency of their interaction. The most simple idea for such is that when two tags co-appear in many queries, their interaction should be recognized by an edge in the tag-tag interaction graph. As illustrated in the right-bottom part of Figure~\ref{fig:classifier}, we use a simple frequency threshold based approach like frequent itemset
mining~\cite{Zaki:2000:Eclat} to build this graph. Using this approach, if two tags co-occur beyond a certain percentage
(known as minimum support in frequent pattern mining literature) of queries in a representative query-set, we build an 
edge between the corresponding tokens in the query. In this manner, through tag association, a query can be converted
to a tag associate graph, where tokens with the associated tags are connected. One can also use other measures besides
frequency, such as, mutual information, for deciding on the edges of the tag association graphs. We found tag association
graph to be sparse, so we also add edges between the adjacent tokens in the query.  Formally speaking, say, $S= s_1, s_2, \cdots, s_M$ is a query and $t_i$ is the tag associated with token $s_i$. If there is a tag pair $(t_i, t_j)$ which is  
frequent (over a representative query dataset) by a frequent pattern mining algorithm, we build an edge between $s_i$ and
$s_j$ in $S$. We also add edges between consecutive tokens in the sentence, i.e.,  $s_i$ is connected with $s_{i\pm 1})$
through an edge. 

\section{Experiment and Result} 

We perform comprehensive experiment to show the effectiveness of \name\ for token classification task. 
Below we first discuss the dataset, and competing methods, followed by experimental results of the experiments.

\subsection{Dataset} \label{dataset} 

\begin{figure}[t]
    \centering
    \includegraphics[scale=0.35]{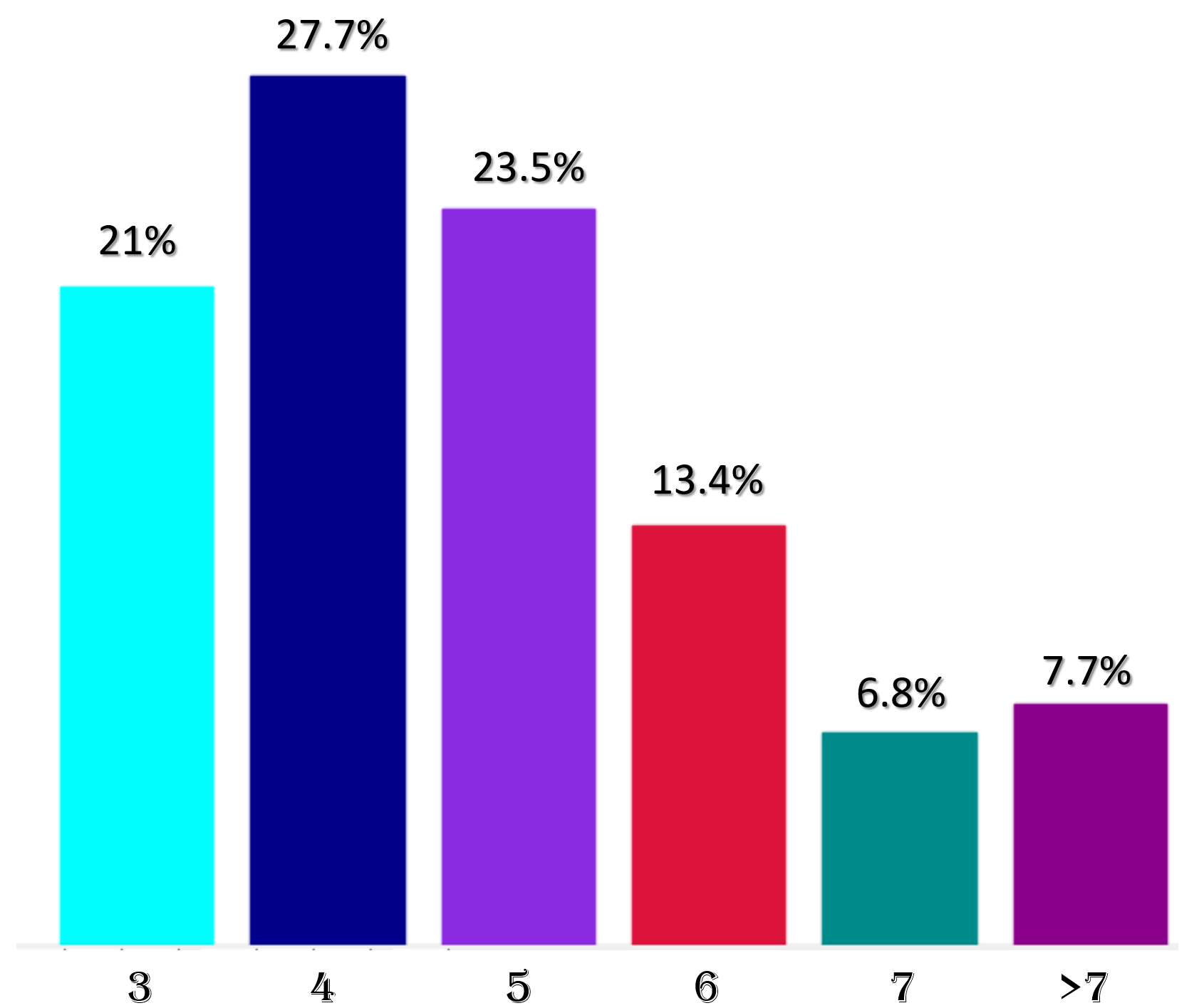}
    \caption{Dataset distribution based on distinct number of words/tokens}
    \label{fig:bar}
\end{figure}

Our dataset was sourced from eBay's query understanding pipeline and comprises train, test, and validation datasets with four columns: source query, destination query, phrases in the source query, and corresponding phrase tags. The train, test, and validation datasets contain 6 millions, 2 millions, and 2 millions instances, respectively, in a 6:2:2 ratio. The distribution of the number of words in a source query is similar across these datasets, as shown in Figure~\ref{fig:bar}. Approximately 28\% of the queries in the dataset are four-word queries, and there are no single-word queries since our focus is on token dropping. Only around 700 queries consist of two words, making them very rare, while queries with three, four, or five words are dominant, accounting for around 70\% of the dataset.

\subsection{Competition Methods} \label{rep}  For comparison, we consider a collection of neural architectures, including
BERT, Sentence-BERT, LLaMA, and Seq2Seq methods. We have also developed several models, such as, eBERT, and 
eBERT-two-tower, by modifying the existing BERT model for our baselines. We discuss all of the competing methods below.

\subsubsection{BERT}
Bidirectional Encoder Representations from Transformers (BERT) is proposed by researchers from Google 
\cite{bert:Embedding}, which is not trained on any specific downstream task but instead on a more generic task called 
Masked Language Modeling. The idea is to leverage huge amounts of unlabeled data to pre-train a model,
which can be fine-tuned to solve different kinds of NLP tasks by adding a task specific layer which maps the 
contextualized token embeddings into the desired output function. In this work we use the pre-trained model \textit{'bert-
base-uncased'} which has a vocabulary of 30K tokens and 768 dimension for each token. We use the BERT embedding for token 
classification.

\subsubsection{eBERT}
eBERT\footnote{https://www.enterpriseai.news/2021/09/15/heres-how-ebay-is-using-optimization-techniques-to-scale-ai-for-its-recommendation-systems/} is an e-commerce specific version of the BERT model. Along with the Wikipedia corpus, 1 billion latest unique item titles are collected to train the model. The eBERT model can represent the e-commerce token better than the original BERt which is trained on Wikipedia corpus. We compare with the pre-trained eBERT model by using
token embedding from this model. We keep the embedding dimension of eBERT similar to BERT which is 768.

\subsubsection{eBERT Two Tower} The BERT and eBERT model we demonstrate only learn from the source query tokens. That's why we design another model which combines the eBERT token embedding with another model which learns the corresponding tag embedding in parallel. In the two tower architecture, the first tower is associated with token embedding and the second tower aims to learns the tag embedding. Finally, both types of representations from both the towers are either concatenated, min pooled, max pooled, mean pooled or combined by the gated architecture in Section~\ref{model}. 
As the two towers models are concatenation of two embedding, each token is represented by $768\times 2$ dimension. 
For convenience, in subsequent discussion this model will be referred as TT.

\begin{table*}
   \caption{Performance of \name\ compared to baseline methods}  
   \setlength{\tabcolsep}{1.9pt}
        \centering
            \renewcommand{\arraystretch}{1.2}
            \scalebox{1}{
                \begin{tabular}{l | c | c| c}
                    \hline
                    \bf Method & \bf F$_1$ Score & \bf Exact-Match & \bf Token-Level\\
                               &  (\% imp.) & {\bf Acc} (\% imp.) & {\bf Acc} (\% imp.) 
                    \\\hline
                    
                     BERT &  0.783 (-)  & 0.291 (-) & 0.684 (-)\\
                     eBERT & 0.807 (3.1) & 0.322 (10.7) & 0.727 (6.3)\\
                    \hline
                    LLaMA &  0.793 (1.3) & 0.31 (6.5) & 0.7 (2.3)\\
                    Sentence-BERT & 0.787 (0.5) & 0.32 (10.0) & 0.697 (1.9)\\
                    \hline
                     eBERT TT + Mean &  0.801 (2.3)& 0.322 (10.7) & 0.727 (6.3)\\
                     eBERT TT + Gated & 0.809 (3.3) & 0.323 (11.0) & 0.728 (6.4)\\
                      \hline
                      Seq2Seq &  0.799 (2.0) &  \bf 0.45 (54.6) & 0.736 (7.6)\\
                      \hline
                       \textsc{QUEEN} &  0.789 (0.7) & 0.334 (15) & 0.729 (6.6)\\\hline
                     \textsc{TagBERT} (Static, Mean) & 0.824 (5.2) & 0.349 (20.0) & 0.748 (9.4)\\
                    \textsc{TagBERT} (Static, Gated) & \bf 0.83 (6.0) & 0.358 (23.0) &   0.757 (10.7)\\
                    \textsc{QUEEN + \name} &  0.827 (5.6) & 0.361 (24.1) & 0.755 (10.4)\\
                    \textbf {\textbf{\textsc{TagBERT (Dynamic)}}} & \bf 0.83 (6.0) & 0.37 (27.0) & \bf 0.76 (11.1)\\\hline
                \end{tabular}
            }

        \label{table:classification-results-80}
\end{table*}

\subsubsection{LLaMA} LLaMA~\cite{touvron2023llama} is another transformer based model developed by researchers in Meta. The model contains 7B to 65B parameters and it is trained on trillions of tokens. The model outperforms GPT 3 and other state of the art methods. The pre-trained model is available online. We use the pretrained model to represent the e-commerce query tokens. The representation of each tokens (4096 dimensional) is learned by LLaMA for token classification.

\subsubsection{Sentence-BERT} Sentence-BERT~\cite{reimers2019sentence} is another pretrained model designed to capture mainly semantic textual similarity. The model uses siamese and triplet network structures to derive semantically meaningful sentence embedding. The model can also be used for token representation for token classification. For 
comparison we use the pretrained weights from the model ``all-MiniLM-L6-v2'' available in Huggingface, which produces
384 dimensinoal vectors for each token.

\subsubsection{Seq2Seq method} We also compare a sequence to sequence model~\cite{liyang:2023} with \name. The model uses pre-trained eBERT architecture for encoding source query. It also has an associated decoder to generate the target query. The generation process continues until an end marker is generated. This model is unaware of tag dependency relation among the words. 

\subsubsection{QUEEN}
We further compare with another model~\cite{Wang2021QUEENNR}, originally designed for query rewriting in the e-commerce domain. While this model is not specifically intended for token classification tasks, we adapt it for this purpose. The model incorporates token embeddings with the original BERT model, where each token is associated with a tag. During tokenization, along with the token ID and positional ID, a tag ID is generated and passed through an embedding layer. The resulting tag embedding is concatenated with the existing BERT wordpiece embedding to align with the existing encoder architecture of BERT. The final feed-forward layer provides embeddings for each token, which are then fed to a softmax classifier layer for token classification. As we are not focusing on query rewriting, we discard the decoder architecture. In our implementation, we experiment with tag embedding dimensions of 16, 32, 64, 128, 256, and 512, ultimately finding the best performance with a dimension of 512.

\subsubsection{TagBERT}
This model is conceptually similar to \name, except for its use of a static graph of dependency relations extracted through frequent pattern mining. Similar to the original \name, for each token, this model learns a 768-dimensional vector.

\subsubsection{TagBERT plus QUEEN}
We propose this hybrid model, which combines the ideas from the QUEEN model with \name. Here, we retain the original dynamic graph architecture while adopting the tokenizer used in the QUEEN model. The BERT architecture is modified to incorporate the tag embedding, while keeping the dynamic graph learning architecture unchanged. Similar to the original QUEEN model, we tune the dimension of the tag embedding to optimize performance.

Table~\ref{table:classification-results-80} provides a comprehensive comparison of performance among all baseline methods, including four variants of TagBERT, namely TagBERT (Mean), TagBERT (Gated), QUEEN plus \name, and \name. In TagBERT (Mean), we simply compute the average of BERT embedding and dependency-aware BERT embedding for each token, while TagBERT (Gated) utilizes the gating mechanism discussed in the Methodology section. Additionally, we present the results for QUEEN, QUEEN plus \name, and all variations of TagBERT and \name as our designed methods, which are separated in the bottom part of the table.  We present $F_1$-score, Exact-Match Accuracy, and Token-Level accuracy of all the models. 
All the evaluation metric values are computed at the token level, except for Exact Match Accuracy, in which prediction
is considered to be accurate only if the predicted phrase is identical to the target phrase.
For the later three metrics, we also show the percent improvement over the corresponding metric value of BERT model.
As can be seen from the table, BERT is the weakest among all the models, which is expected as this model is not trained
using an eCommerce queries, which have distinct characteristics such as being shorter, aspect-dominant, and lacking in dependency structure. However, when we update the BERT training corpus by adding e-commerce queries and categories,
the performance improves, as can be seen by observing the results of eBERT. However, among all the models,
different variants of \name\ are the best considering every metric, except Exact-Match Accuracy, in which the 
Seq2Seq model performs the best and \name\ is the second best. For example, the $F_1$-score, Exact-Match Accuracy,
and Token-Level accuracy of \name\ are 0.83, 0.37, and 0.76, which are 6\%, 27\%, and 11.1\% better over the BERT
model.
In this table we have shown two variants of static TagBERT (mean and gated). In our experiments, we actually combined the BERT 
and tag-dependency-aware BERT representations using different aggregation functions, such as mean, max, min and gated
pooling, and the results from the best two performers are shown in this table. However, the gated version outperforms 
all others, because this version makes an optimal combination of BERT and tag-dependency-aware BERT embedding by
optimizing the gate probability based on the relative quality of these two embeddings.  As a result, the performance of 
\name (Dynamic) is shown for the gated version only, as it surpasses the performance of the other aggregation methods.

Clearly, tag-interaction aware attention mechanism makes \name\ superior to eBERT. Can same improvement be achieved
by simply combing eBERT embedding with tag embedding? This can be answered by comparing \name\ with eBERT TT; the latter
combines eBERT token embedding
and tag embedding to obtain a representation of a token. Our results show that percent improve of \name\ over BERT is
around twice of the percent improvement of eBERT TT over BERT. This shows that attention over tag interaction which
\name\ utilizes yields superior signal that the tag embedding for solving token classification task.

Among the other models, LLaMA and Sentence-BERT performs better than BERT in terms of all evaluation metrics. However, these models are not e-commerce specific. So like BERT, these models do not even outperform eBERT, let anone \name. 
Among LLaMA and Sentence-BERT, LLaMA performs better. This is expected as LLaMA is a larger language model with billions 
of parameters. Furthermore, as we compare \name\ with Seq2Seq model, the latter wins in Exact-Match accuracy. This is 
expected as the Seq2Seq model is trained to generate the target query so its loss function favors exact matching, 
whereas \name's loss function favors token level classifcation. On the other hand, the QUEEN model exhibits comparatively less appealing performance. It is worth noting that the QUEEN model directly learns from the sequence of tags, which may not effectively capture the dependency relation. In contrast, all versions of TagBERT incorporate the dependency relation either as a static or dynamic graph, which appears to play a crucial role in TagBERT's performance.

\subsection{Results} 
\begin{table}
   \caption{Performance of \name\ for different query length in terms of word count}  
   \setlength{\tabcolsep}{1.9pt}
        \centering
            \renewcommand{\arraystretch}{1.2}
            \scalebox{1}{
                \begin{tabular}{l | c | c| c }
                    \hline
                    \bf Length & \bf F$_1$ Score & \bf Exact Match Acc& \bf Token Level Acc \\\hline
                     3 & 0.85 & 0.54 & 0.78\\
                     4 & 0.83 & 0.37 & 0.75\\
                     5 &  0.82 & 0.29 & 0.74\\
                     6 &  0.82 & 0.23 & 0.74\\
                     7 &  0.83 & 0.2 & 0.75\\
                      \hline
                \end{tabular}
            }
        \label{table:length-performance}
\end{table}

In the last section of Table 1, we showcase the performance of different variants of TagBERT. Overall, these methods exhibit similar performance, but \name (Dynamic) achieves the highest scores. The reason for this superiority lies in the fact that among all the variants, only \name (Dynamic) learns the graph dynamically. While our static graph learning method also performs quite well, the dynamic graph learning approach proves to be more effective in learning the edge probabilities. Admittedly, in some cases, the probability values can introduce noise since there must be some probability between any pair of tokens. Nevertheless, the superior performance of \name\ underscores the advantages of the dynamic graph learning method.
Besides training data form where the static tag-interaction graph is built can be outdated, whereas a dynamic graph based method does not suffer from this limitation.
We conduct an additional experiment using \name, varying the number of words in a source query. The findings of this experiment are presented in Table~\ref{table:length-performance}. When it comes to token-level performance, including precision, recall, F$_1$ Score, and Token Level Accuracy, there is minimal change observed when the length of the query is increased. However, the accuracy of exact matches decreases as the number of words in a query increases, as it relies on the predicted query being an exact match to the target query.

\section{Conclusion} 
In this paper, we propose \name, a tag-interaction aware transformer model for identifying important tokens for the task 
of query reformulation. Through \name, we show that by considering attention only along the tag-interaction edges, an
auxiliary token embedding can be obtained, which when combined with a BERT-like embedding through gated mechanism,
improves the token classification performance substantially. Experiments with a large eCommerce dataset from eBay 
validates the superiority of \name, over six baseline models using three evaluation metrics.

\section{Acknowledgement}
 This project is funded by a grant by eRUPT program of eBay. We also acknowledge the support of eBERT from CoreAI.

\nocite{Feigenbaum1963a}
\nocite{Hill1983a}
\nocite{OhlssonLangley1985a}
\nocite{Matlock2001}
\nocite{NewellSimon1972a}
\nocite{ShragerLangley1990a}

\printbibliography

\end{document}